\documentclass[a4paper,fleqn]{cas-dc}
\usepackage{placeins}

\usepackage[sort,numbers]{natbib}

\def\tsc#1{\csdef{#1}{\textsc{\lowercase{#1}}\xspace}}
\tsc{WGM}
\tsc{QE}
\tsc{EP}
\tsc{PMS}
\tsc{BEC}
\tsc{DE}

\usepackage{graphicx}
\usepackage{lscape}
\usepackage{array}
\usepackage{graphicx}
\usepackage{multicol}
\usepackage{multirow}
\usepackage{amssymb}
\usepackage{pifont}
\usepackage{tikz}
\usepackage{verbatim}
\usepackage{array}
\usepackage{longtable}
\usepackage{supertabular}
\usepackage{url}
\usepackage{comment}
\usepackage[labelformat=simple]{subcaption}

\DeclareCaptionLabelFormat{subcaptionlabel}{\normalfont(\textbf{#2}\normalfont)}
\begin{document}
\shorttitle{Explainable Artificial Intelligence for Drug Discovery}
\shortauthors{Roohallah et~al.}

\title [mode = title]{Explainable Artificial Intelligence for Drug Discovery and Development - A Comprehensive Survey}                      
\author[1]{Roohallah Alizadehsani}[type=editor,
                        ]
\ead{r.alizadehsani@deakin.edu.au}


\address[1]{Institute for Intelligent Systems Research and Innovation (IISRI), Deakin University, Geelong, VIC 3216, Australia}

\author[2]{Solomon Sunday Oyelere}[type=editor,
                        ]
\cormark[2]                      
\address[2]{Department of Computer Science, Electrical and Space Engineering, Luleå University of Technology, Skellefteå, Sweden }
\ead{solomon.oyelere@ltu.se}

\author[3]{Sadiq~Hussain}[type=editor,
                        ]

\address[3]{Department of Statistics, Dibrugarh University, Dibrugarh 786004}
\ead{jiten_stats@dibru.ac.in}

\author[4]{Rene~Ripardo~Calixto}[type=editor,
                        ]

\address[4]{Department of Teleinformatics Engineering, Federal University of Ceará, Fortaleza 60455-970, Brazil}
\ead{reneripardo@gmail.com}

\author[4]{Victor~Hugo~C.~de~Albuquerque}[type=editor,
                        ]

\address[4]{Department of Teleinformatics Engineering, Federal University of Ceará, Fortaleza 60455-970, Brazil}
\ead{victor.albuquerque@ieee.org}

\author[5]{Mohamad Roshanzamir}[type=editor,
                        ]

\address[5]{Department of Computer Engineering, Faculty of Engineering, Fasa University, Vali asr Blvd, Fasa 74617-81189, Iran}

\author[6]{Mohamed~Rahouti}[type=editor,
                        ]

\address[6]{Department of Computer \& Information Science, Fordham University, Bronx, NY 10458, USA}
\ead{mrahouti@fordham.edu}

\author[7]{Senthil Kumar Jagatheesaperumal}[type=editor]
\ead{senthilkumarj@mepcoeng.ac.in}
\address[7]{Department of Electronics and Communication Engineering, Mepco Schlenk Engineering College, Sivakasi, Tamil Nadu, India}

\cortext[cor2]{Corresponding author}



\begin{abstract}
The field of drug discovery has experienced a remarkable transformation with the advent of artificial intelligence (AI) and machine learning (ML) technologies. However, as these AI and ML models are becoming more complex, there is a growing need for transparency and interpretability of the models. Explainable Artificial Intelligence (XAI) is a novel approach that addresses this issue and provides a more interpretable understanding of the predictions made by machine learning models. In recent years, there has been an increasing interest in the application of XAI techniques to drug discovery. This review article provides a comprehensive overview of the current state-of-the-art in XAI for drug discovery, including various XAI methods, their application in drug discovery, and the challenges and limitations of XAI techniques in drug discovery. The article also covers the application of XAI in drug discovery, including target identification, compound design, and toxicity prediction. Furthermore, the article suggests potential future research directions for the application of XAI in drug discovery. The aim of this review article is to provide a comprehensive understanding of the current state of XAI in drug discovery and its potential to transform the field. 
\end{abstract}




\maketitle
\section{Introduction} 
\label{sec:intro}
The field of drug discovery and development has started playing a significant role in the healthcare industry, in order to find new compounds and therapeutic targets that can successfully cure a variety of diseases~\cite{ekins2019exploiting}. The traditional drug development method has been laborious, and resource-intensive, and experienced major difficulties in identifying promising therapeutic candidates over the previous few decades. However, the landscape of drug development has undergone a radical upheaval since the advent of artificial intelligence (AI) and machine learning (ML) technologies, which promise to quicken and improve the process~\cite{lavecchia2019deep}.

AI and ML models have shown the ability to analyze enormous datasets, discover insightful patterns, and generate predictions for identifying prospective drug candidates and targets. Lead optimization~\cite{ferreira2019admet}, virtual screening~\cite{maia2020structure}, compound design~\cite{schneider2020rethinking}, and medication repurposing~\cite{sahoo2021drug} are just a few of the domains where the use of AI and ML in drug development has already produced promising outcomes. These models have the potential to greatly overshoot the success rate of drug discovery and decrease the time and expense involved as they become more complex and potent. As AI and ML models become more complicated, a major issue in the area has emerged, which is the lack of transparency and interpretability~\cite{elbadawi2021advanced}. Although these models have excellent predictive powers, it still remains challenging to explain why they make such predictions. This lack of interpretability can make it more challenging for researchers, doctors, and regulatory bodies to trust and accept AI and ML-driven predictions in the context of drug discovery. Validating and ranking the discovered targets or compounds also becomes hectic without knowledge of how AI algorithms make decisions~\cite{koivisto2022advances}.

Such aforementioned interpretability gap could be addressed by Explainable Artificial Intelligence (XAI), which tries to offer clear and intelligible justifications for the predictions made by AI and ML models~\cite{kirbouga2023explainability}. In addition to increasing trust and acceptance, XAI also makes it possible for researchers to spot any biases, inaccuracies, or limits in the underlying data or model architecture by making it possible for humans to understand the logic behind the model's predictions.

While the integration of XAI approaches to improve interpretability is still an emerging area of research, the application of AI and ML in drug discovery is quickly progressing. Comprehensive studies that explore the state-of-the-art in XAI for drug development are severely lacking. It is challenging to gain a comprehensive knowledge of the state of the field and its potential because existing research frequently focuses on particular XAI features or unique drug discovery tasks. Additionally, the majority of AI-driven drug development research stresses predictive performance over interpretability~\cite{kurosaki2022development}, frequently forgoing the latter in favor of greater accuracy. This imbalance is critical since regulatory approval and real-world use of AI-driven drug development methods depend more and more on interpretability. To ensure safe, effective, and moral medication development, researchers, physicians, and policymakers need a comprehensive knowledge of how AI algorithms generate their predictions~\cite{sarkar2023artificial}.

The proposed article offers a complete and organized overview of the present state-of-the-art in XAI for drug discovery, that aims to fill the vacuum in the existing literature. The article will emphasize the benefits, drawbacks, and prospective uses of the various XAI approaches used in drug development by synthesizing and analyzing them. The thorough examination will not only go over the various XAI techniques but also show how they may be used for various drug discovery processes, such as target identification, chemical design, and toxicity prediction. Some of the significant contributions of the article are as follows:

\begin{itemize}
    \item XAI and drug discovery methods fundamentals and their significant role in promoting the healthcare sector.
    \item Detailed exploration of XAI applications in healthcare, focusing on drug discovery tasks.
    \item Evaluation of XAI frameworks' strengths and limitations in drug discovery considering performance parameters.
    \item Critical discussion of lessons learned, limitations, and research challenges in implementing XAI for drug discovery.
    \item Identification of future research directions, inspiring innovative approaches for XAI in drug discovery.
\end{itemize}

\section{Background}

\subsection{XAI (What is XAI?)}
Since the initial theoretical studies on AI, there have been a large number of innovative applications using AI in order to help in the development of society. An AI system must perform decision-making in a reliable, secure way that can be audited by a user with little technical knowledge. For the user to understand how the AI makes decisions, techniques are needed that seek to explain why certain AI input information produces a specific result. To do this task, XAI is used.

\begin{figure}[th!]
       \begin{subfigure}[b]{1\columnwidth}
       \captionsetup{justification=centering}
         \centering
         \includegraphics[width=0.8\columnwidth]{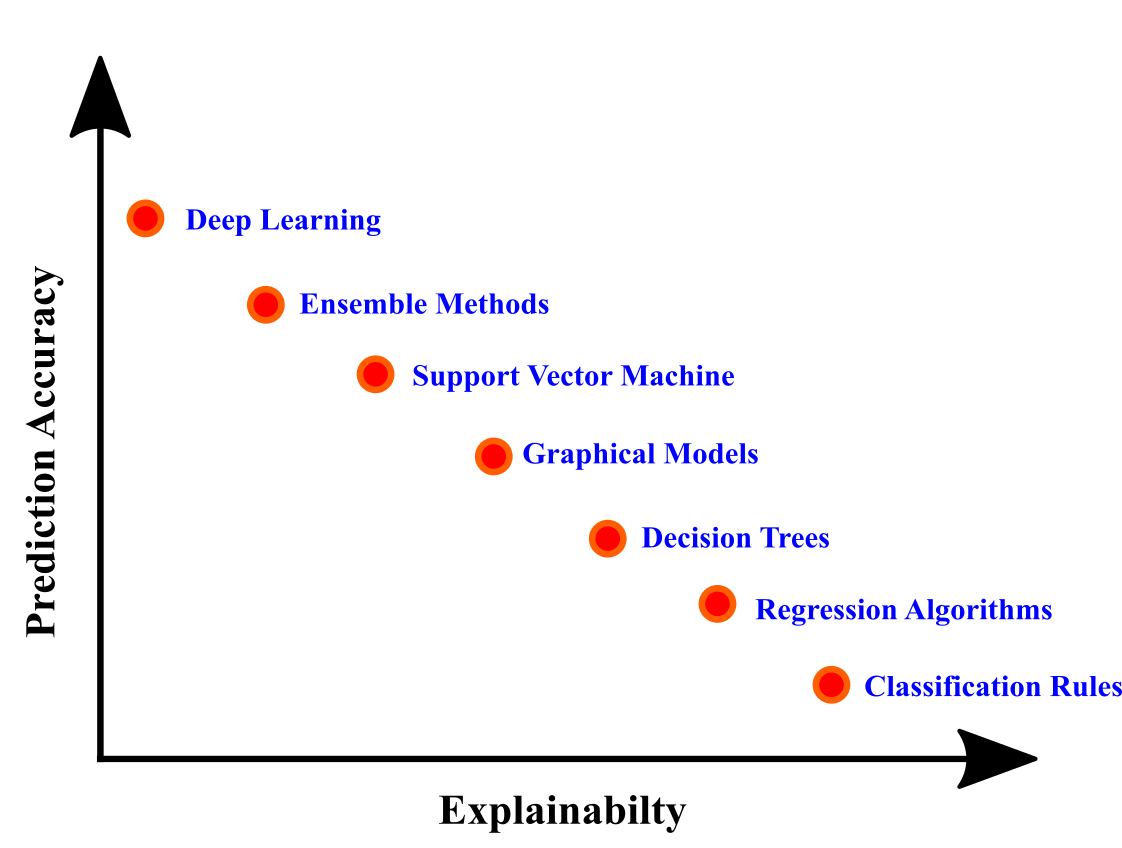}
         \caption{Explainability vs. Prediction Accuracy}
         \label{fig:xaipred}
     \end{subfigure}
     \hfill
     \hspace{0.01\textwidth}
     \begin{subfigure}[b]{1\columnwidth}
     \captionsetup{justification=centering}
         \centering
         \includegraphics[width=0.8\textwidth]{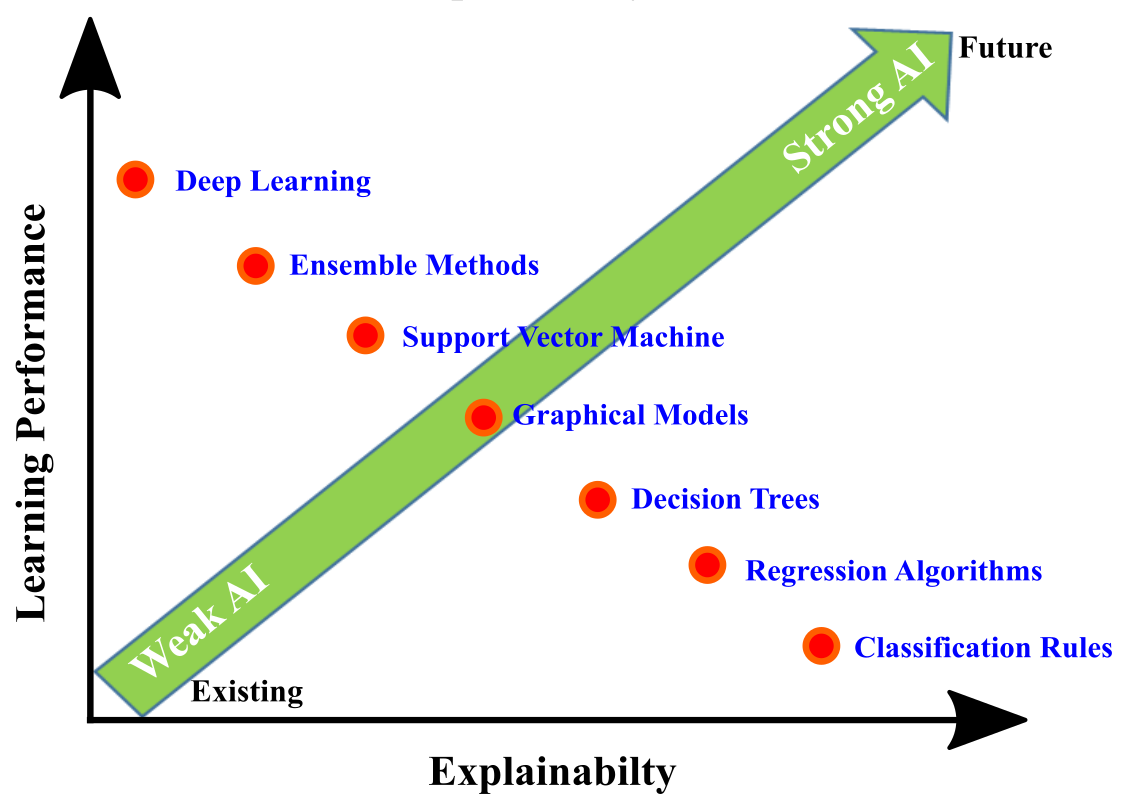}
         \caption{Explainability vs. Learning Performance}
         \label{fig:xaiweak}
     \end{subfigure}
         \caption{Comparison of popular XAI models based on the accuracy and performance metrics on explanability}
     \label{fig:xaiperf}
\end{figure}

XAI is a technique that aims to logically explain to a user the behavior and decision-making of an AI system~\cite{haque2023surveixai}. XAI evolved due to the need to interpret decisions made by a machine learning model, in this case, XAI indicates the logic performed by a model to reach conclusions from a classification process. Therefore, XAI provides corrective measures, error prediction, and explanation of failures that occur in a system. This allows confidence in the model results~\cite{Aboutorab2021survey}. Fig.~\ref{fig:xaiperf} shows the comparison of popular XAI models based on the accuracy and performance metrics of their explainability characteristics. On the path of marching towards stronger AI from weaker AI, robust XAI models are in demand for sensitive healthcare applications like cancer treatment, drug delivery, and drug discovery. 

\subsection{Drug Discovery Methods}
Drug discovery is an area that develops drugs that eliminate or minimize disease. According to Pandiyan and Wang, ~\cite{sanjeevi2022comprehensive}, the average period of research and development lasts around 10 to 17 years for the development of a new form of medicine and with an estimated average value of more than US \$ 2.8 billion. Other challenges are quite present in these researches, such as the low efficiency and high cost of conventional methods to carry out drug discoveries~\cite{wei2023artificialIntelligence}. So, there is a need to develop new methods that can minimize these problems. An alternative to minimize these problems in drug discovery can be through AI. Through AI in drug discovery, applications are possible for virtual screening, drug design, reaction prediction, protein design, and other predictive tasks ~\cite{deng2022artificial}.

Singh et al.~\cite{vishakha2023multiscale} used multi-scale temporal convolutional networks (MSTCN) to develop a model based on deep learning named MSTCN-ABPpred (BL) that classifies antibacterial peptides (ABPs). The main contribution of this study was the continuous learning capacity that the model has in the face of new data for re-training.

\subsection{Search Strategy}

The search strategy used in this article to find scientific papers is divided into three stages. The first step is to come up with terms related to Explainable AI and Drug Discovery. Therefore, the main search terms were: “Explainable AI” and “drug discovery”. Other terms were also used, such as drug development, drug design, drug toxicity, Artificial Intelligence, Machine Learning and Neural Networks. In the second stage, the search was carried out in journal articles. Duplicate articles were excluded.

In the third stage, the insertion criteria were: works with the aim of drug design, reaction prediction, protein design, target identification, compound design and toxicity prediction through AI, ML and especially XIA. This search selects works published since 2019 in order to find recent research. Studies were searched in IEEE Xplore, ScienceDirect, Springer, PLOS ONE, Inderscience, MDPI, Hindawi, Wiley and peerJ databases. 
        



\section{XAI in Healthcare} 
\label{sec:xaihealth}

\subsection{Drug discovery and XAI}
In the past 20 years, Computer Aided Drug Discovery (CADD) proved its efficacy and its significance has been improved in leaps and bounds. In recent years, many Artificial Intelligence in Drug Discovery (AIDD) strategies were utilized for drug discovery. Sharma et al.~\cite{sharmaartificial} discussed the pros and cons of AIDD in their survey. At the outset, AI was employed to design logical programming platforms (LISP, Prolog) at par with usual programming architectures. Later, as a sub-topic of machine learning (ML), many new techniques in Knowledge Base Systems (KBS) like Genetic Algorithms (GA), Support Vector Machines (SVM), Artificial Neural Networks (ANN), Fuzzy Systems (FS), pattern recognition tools, and Deep Learning (DL) were devised that were utilized in AIDD. The overlap between the operations performed at the atomic level by molecular modeling approaches and AIDD-based operations is enhancing with tremendous growth. There are lots of efficient AI-based methods in drug discovery, but their applications are limited in both functionality and capability. One of the major limitations of AI methods such as neural networks is that they are often regarded as black boxes. As these methods depend on the training dataset, there is always a concern for generalizing a situation that was not depicted in the dataset. One of the drawbacks of the genetic algorithm techniques is that there is no surety of achieving optimal solutions. When the data size is not large enough, the performance of the deep learning models deteriorates. Sufficient reliable data is the key to the success of AIDD models. 

Sahoo et al.~\cite{sahoo2021comprehensive}reviewed how AI can aid in finding the drug-like stuff in the compound screening phase predicting ADMET and the Structure-Activity Relationship (SAR) in lead detection and optimization stages, sustainable development of chemicals in the synthesis stages up to the assistance of AI in the conduct of repurposing and clinical trials. There is a paradigm shift from hit and trial approach in drug discovery to traditional drug discovery and development. To overcome the black-box approach of the AI methods, XAI is the flag bearer. Machines or robots can compile the data, especially in the rational drug discovery process so that drug designers plan better and the synthesis process is easier.

XAI is an area that explains predictions made by AI models. Gillani et al.~\cite{gillani2022explainable} defined the value of features while predictions were made. The prediction of cancer’s reaction to a particular treatment or drug efficiency is a hot topic. In drug discovery, based on huge genomics data, drug sensitivity forecasting is a vigorous process. On the other hand, drug personalization is an arduous and tedious matter. XAI imparts dependency and confidence. Their research was a step towards the understanding of drug chemical structures and deep learning strategies on gene expression. 

Non-Small Cell Lung Cancer (NSCLC) demonstrates inherent heterogeneity at the molecular level that helps in discriminating between two of its subtypes - Lung Squamous Cell Carcinoma (LUSC) and Lung Adenocarcinoma (LUAD). Dwivedi et al.~\cite{dwivedi2023explainable}presented a new XAI-based deep-learning approach to locate a small set of NSCLC biomarkers. Their framework employed an Autoencoder to reduce the input feature space and NSCLC instances were classified into LUSC and LUAD using a feed-forward neural network. They detected that 14 of the biomarkers are druggable. The survivability of the patients could be predicted by 28 biomarkers. They observed that seven of the newly discovered biomarkers had never been utilized for NSCLC subtyping and could be promising for the targeted therapy of lung cancer.

Due to a lack of prior knowledge, the development of XAI techniques poses a challenge to the quantitative assessment of interpretability. Rao et al.~\cite{rao2022quantitative} devised five molecular benchmarks to exploit quantitatively XAI techniques utilized in Graph Neural Networks (GNN) approaches and compared them with human experts. The vital substructures for chemists could be identified by the XAI techniques that delivered informative and reliable explanations and could be empirically demonstrated. 

The cost and time of new drug development can be mitigated as drug repositioning exhibits great potential. Drug repositioning can confirm the necessity of pharmacological effects on biomolecules for application to new diseases by omitting different R \& D processes. In a disease-drug association prediction architecture, biomedical explainability provides insights into ensuing in-depth studies. Takagi et al.~\cite{takagi2022graphix} introduced an explainable drug re-positioning approach called GraphIX utilizing biological networks and examined its explainability quantitatively. They applied a graph neural network to learn node features and network weights. 

In drug discovery, deep learning can play a significant role if it imparts in development efforts and experimental research~\cite{bajorath2022artificial}. This will require time and effort as it needs long drug development times. Medicinal chemists witnessed state-of-the-art methodological advances via DL in synthesis design but lack in DL tools in compound optimization. It is a challenging task to transform expert-dependant DNN architectures into widely usable and robust compound models. There is a need of narrowing the gap between experiments and DL for enhancing the confidence of practitioners to generate trust in the predictions. 

Most XAI techniques do not come as readily operational, ‘out-of-the-box’ solutions, which need to be configured for each application~\cite{jimenez2020drug}. Additionally, deep knowledge of the problem domain is vital to detect which model decisions demand further explanations, which types of answers are meaningful to the user, and which are instead expected or trivial. Finding such types of solutions needs the collaborative effort of biologists, chemists, data scientists, chemoinformatics, and deep-learning experts to confirm that XAI methods deliver trustworthy answers and serve the intended purpose. XAI in drug discovery suffers a dearth of an open-community platform for enhancing model interpretations, and software and sharing training data. For federated, decentralized model deployment and secure data handling across pharmaceutical companies, MELLODDY (Machine Learning Ledger Orchestration for Drug Discovery, melloddy.eu) is a great initiative. This kind of collaboration fosters acceptance, validation, and development of XAI. 

Karger et al.~\cite{karger2022using} conducted a bibliometric analysis of AI for drug discovery. They considered 3884 articles published between 1991 and 2022. They surveyed the most productive countries, institutions, and funding sponsors in the domain. They utilized thematic analysis and science mapping to identify the thematic areas and core topics. They also outlined future research avenues in the field of drug discovery using AI. The findings indicated the multidisciplinary nature of AI and its understanding of discovering drugs. They encouraged the utilization of unsupervised learning algorithms to identify patterns in unlabelled data to address unknown drug discovery problems. They considered the explainability of AI algorithms as another future research need. 

Askr et al.~\cite{askr2022deep} performed a systematic Literature review (SLR) that integrated the current DL strategies along with various types of drug discovery problems, drug–drug similarity interactions (DDIs), Drug–target interactions (DTIs), drug-side effect predictions and drug sensitivity and responsiveness. They are linked with the benchmark databases and data sets. They also discussed related topics like digital twining (DT) and XAI and how they support drug discovery issues. Moreover, success stories of drug dosing optimization were also narrated. They presented open problems for future research challenges as well. 

Real World Data (RWD) and AI showcased their potential but applied in limited areas across several phases of the drug development process. Most of the AI studies aimed at the detection of adverse events (AE) from clinical narratives in Electronic Health Records (EHRs) and a few of them examined clinical drug repurposing and trial recruitment optimization. The AI techniques on RWD demonstrated its efficacy by generating novel hypotheses and exploring previously unknown associations. Nevertheless, challenges and knowledge gaps still exist, for example, the difficulty of sharing clinical data, data quality issues, and lack of transportability and interpretability in AI models. Chen et al.~\cite{chen2021applications} surveyed the latest advancements of AI in drug discovery and their challenges. Enhancing the capability of DL models that could handle heterogeneous and longitudinal RWD and new research opportunities in drug development are some of the areas to exploit. 

A huge amount of resources, capital, and time are required to search for effective treatment of existing and novel diseases. The dearth of antimicrobial agents for the treatment of emerging infectious diseases like COVID-19 is a major concern. AI and other in silico techniques can boost the drug discovery arena by proving more cost-friendly approaches with better clinical tolerance. Numerous researchers have been working on devising AI platforms for hit identification, lead optimization, and lead generation. In~\cite{bess2022artificial}, Bess et al. investigated the effective AI techniques that revolutionized the pharmaceutical sciences and drug discovery.  

Han et al.~\cite{han2021challenges} described the challenges of XAI in biomedical science. AI systems should not only have good results but they should yield good interpretability. Good explainability is missing in the prevailing AI models in bio-medical sciences and hence hinders creating transparency and trustworthiness.  Reliable results with good interpretations of why it works rather than only it works is the key and there is an urgent need to design XAI models in the domain. In biomedical data science, numerous types of data range from EMRs, bio-image data, text, high-dimensional omics data, and sequencing data. The complexity, nonlinearity, and size of the data along with problems mostly disease-oriented force the AI methods to make the trade-off between good explainability and good performance. The learning biases created by AI methods are another issue that prevents the techniques from showcasing the minimum interpretations. The learning biases can be the results of imbalanced data, wrong parameter setting or tuning, mismatched interactions between a certain type of data and AI methods, or other complicated problems that may not be identified by biomedical data scientists. There is remarkable progress in XAI recently where knowledge-based data representation, learning process visualization, rule-based learning, and human-centered AI model evaluation were exploited to improve AI explainability. 

By extracting features and relations, performing reasoning, and structuring information, knowledge graphs play a vital role in XAI for explainability. Rajabi et al.~\cite{rajabi2022knowledge}focused on the role of knowledge graphs in XAI models in healthcare. Based on their review, they asserted that knowledge graphs in XAI may be utilized for the detection of adverse drug reactions, drug-drug interactions, and healthcare misinformation and to mitigate the research gap between AI-based models and healthcare experts. They also pointed out how to leverage knowledge graphs in post-model, in-model, and pre-model XAI models in healthcare to enhance their explainability.

\section{Impact of XAI on Drug Discovery and Development} 
\label{sec:impactxai}

\subsection{AI in Drug Discovery}
The field of AI in drug discovery is a rapidly growing and evolving area of research, that intends to enhance the efficiency and accuracy of the drug discovery process. In recent years, AI has been applied in several stages of drug discovery, including target identification, molecular screening, lead optimization, and toxicity prediction. One of the key outcomes of AI in drug discovery is the ability to analyze and process vast amounts of biological data, including genomic, proteomic, and pharmacological data, to identify potential drug targets~\cite{paul2021artificial}. Additionally, AI algorithms can be used to screen virtual libraries of chemical compounds to identify potential drug candidates that fit specific criteria. Another significant outcome is the ability of AI to predict drug toxicity and safety more accurately compared to traditional methods~\cite{basile2019artificial}. AI models can analyze large amounts of data on chemical compounds and their interactions with biological systems, providing a more comprehensive understanding of their potential side effects. Moreover, AI can also be used to optimize lead compounds to improve their efficacy, pharmacokinetics, and drug-like properties, allowing for the development of better drugs with fewer side effects~\cite{kumar2022decade}.

\subsection{Role of XAI in Drug Discovery}
\begin{figure*}[ht!]
  \centering \includegraphics[width=\textwidth]{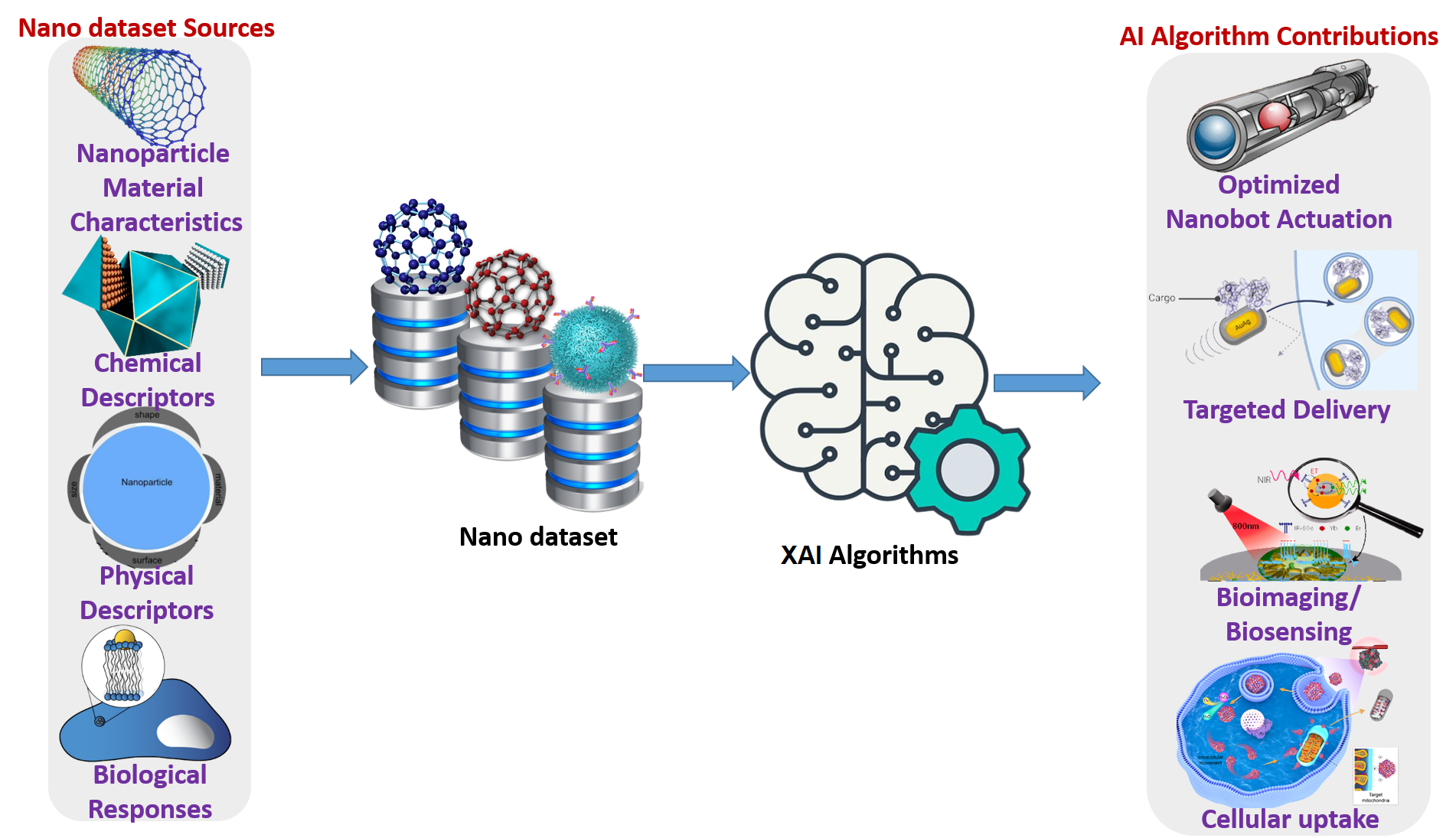}
  \caption{Role of XAI in Drug Discovery and other contributions towards Drug Delivery}
   \label{fig:method}
\end{figure*}

AI has shown great promise in accelerating the drug discovery process and improving the efficiency and accuracy of various stages of drug development. However, further research is required to validate and optimize AI models for drug discovery, as well as to address ethical and regulatory challenges~\cite{iswarya2022drug}. XAI is an important area of research in drug discovery that focuses on developing AI models that can provide clear and transparent explanations for their predictions and decisions. Fig.~\ref{fig:method} shows the schematic representation on the impact of XAI approaches on drug discovery applications.

In drug discovery, XAI can play a crucial role in ensuring the transparency and accountability of AI models, particularly in critical decision-making processes such as lead optimization and toxicity prediction~\cite{rao2022quantitative}. This can help to build trust and confidence in the outcomes generated by AI models, and facilitate the adoption of AI technologies in the pharmaceutical industry. XAI can also help to identify and mitigate potential biases in AI models, ensuring that they produce unbiased and fair predictions. This is particularly important in drug discovery, where biases can have serious consequences, including the development of ineffective or harmful drugs~\cite{jimenez2020drug}. Furthermore, XAI can help to improve the interpretability of AI models, allowing researchers to understand how AI models are making predictions and to identify potential flaws or limitations in their algorithms. This can facilitate the development of more accurate and reliable AI models for drug discovery~\cite{vo2022road}. Furthermore, it also helps to ensure the development of safe and effective drugs. Some of the major roles XAI can play and impact to revolutionize the  drug discovery and development process are:

\begin{itemize}
    \item \textit{Data Analysis:} XAI algorithms can help in the analysis of large amounts of complex and diverse data, including chemical, biological, and clinical data, to identify potential drug targets, predict drug efficacy and toxicity, and optimize drug designs~\cite{harren2022interpretation}. Data analysis in XAI drug discovery involves the use of advanced computational methods and machine learning algorithms to process and analyze vast amounts of data generated from various sources in the drug discovery process. This data can include molecular structures, biochemical assays, and high-throughput screening data, as well as pre-clinical and clinical trial data.

    \item \textit{Decisions Based on Evidence:} Through XAI models we can positively expect interpretable and transparent reasoning behind their predictions and decisions, enabling researchers and regulators to better understand and evaluate the evidence behind the predictions~\cite{antoniadi2021current}.  In drug discovery, decisions are based on evidence that is generated by AI algorithms. These algorithms analyze large amounts of data from various sources such as medical records, scientific publications, and experimental results to identify potential drug candidates.

    \item \textit{Improved Clinical Trial Design:} XAI can help to identify the most appropriate patient population for clinical trials and improve the design of clinical trials by predicting the likelihood of success and identifying potential adverse effects. It is a key aspect of XAI  drug discovery, through which they are used to test the safety and efficacy of new drugs in humans~\cite{wang2022extending}. XAI can be used to improve the clinical trial design in predictive modeling, patient selection, and safety measures

    \item \textit{Personalized Medicine:} To analyze patient data and predict the response of individuals to specific treatments, and enable the development of more personalized and effective treatments XAI algorithms are largely helpful~\cite{shen2021fourth}. XAI drug discovery can play an important role in enabling personalized medicine by using AI algorithms to analyze large amounts of data to support evidence-based decision-making through drug repurposing and real-time monitoring~\cite{drance2022neuro}.

    \item \textit{Improved Regulation:} XAI can help regulators to make more informed decisions by providing interpretable and transparent reasoning behind the predictions made by AI models, ensuring that drugs are developed and approved safely and effectively~\cite{jimenez2020drug}. The goal of improved regulation is to ensure that new drugs are safe and effective for use in humans, while also promoting innovation in the drug discovery process. Through risk assessment strategies, this information can then be used to inform regulatory decisions and ensure that the new drugs are safe for use in humans.
\end{itemize}

As XAI has the potential to significantly improve the efficiency, speed, and accuracy of drug discovery and development, it ultimately leads to the development of more effective and safer drugs for patients.

\subsubsection{Transparency and accountability}

XAI models can provide clear explanations for their decisions and predictions, making it possible for researchers to understand how they arrived at their outcomes. In drug discovery, XAI models can help to address critical issues such as lead optimization and toxicity prediction~\cite{sharma2022fairness}. They can provide a deeper understanding of the underlying biological mechanisms of drugs, allowing researchers to make informed decisions about their development.

Moreover, XAI models can also help to mitigate potential biases in AI algorithms, which can have serious consequences in drug discovery, including the development of ineffective or harmful drugs. This can help to ensure that AI models produce unbiased and fair predictions, regardless of demographic or socio-economic factors~\cite{islam2022hgsorf}. Further, XAI models can improve the interpretability of AI algorithms, allowing researchers to identify potential limitations or flaws in the algorithms and make improvements. This can lead to the development of more accurate and reliable AI models for drug discovery~\cite{jimenez2020drug}. The development of transparent and accountable XAI models is crucial for ensuring the development of safe and effective drugs, providing researchers with a deeper understanding of the underlying biological mechanisms and improving the accuracy and reliability of AI algorithms in drug discovery~\cite{askr2022deep}.

\subsubsection{Bias management}

Mitigating potential biases in Explainable Artificial Intelligence (XAI) models for drug discovery is crucial for ensuring that AI algorithms produce fair, unbiased, and trustworthy predictions. Bias in AI algorithms can have serious consequences in drug discovery, including the development of ineffective or harmful drugs~\cite{chen2021applications}. There are several methods that can be used to mitigate biases in XAI models for drug discovery:

\begin{itemize}
    \item \textit{Data pre-processing:} This involves ensuring that the training data used to develop XAI models is representative of the population and does not contain any biases~\cite{werder2022establishing}. This can be done by using diverse and balanced datasets and removing any irrelevant or biased features from the data.

    \item \textit{Model selection:} Choosing appropriate XAI algorithms that have been shown to reduce biases is another important step. For example, some algorithms, such as decision trees and random forests, have been shown to be less prone to biases compared to other algorithms~\cite{antoniadi2021current}.

    \item \textit{Model interpretation:} XAI models that provide clear and transparent explanations for their predictions and decisions can help to identify potential biases in the algorithms~\cite{askr2022deep}. This can be done by analyzing the model's feature importance or decision paths and comparing them to expert knowledge.

    \item \textit{Model validation:} Evaluating XAI models on independent datasets can help to identify and mitigate biases in the algorithms. This can be done by comparing the predictions made by the model with ground truth data and identifying any discrepancies~\cite{antoniadi2021current}.

    \item \textit{Human oversight:} It is crucial in ensuring that XAI models are fair and unbiased, which can involve domain experts in the development and validation of XAI models, and regularly review the models' outcomes to identify and mitigate any biases~\cite{vale2022explainable}.

\end{itemize}

Ultimately, mitigating biases in XAI models for drug discovery is crucial for ensuring the development of safe and effective drugs. By using a combination of these methods, it is possible to reduce the potential for biases in XAI algorithms and to improve the accuracy and reliability of AI models in drug discovery.

\subsubsection{Interpretability concerns}
Interpretability is one of the key aspects of XAI models for drug discovery, as it allows researchers to understand how AI algorithms are making predictions and decisions~\cite{jimenez2020drug}. This can provide important insights into the underlying biological mechanisms of drugs and improve the accuracy and reliability of AI models. There are several methods that can be used to improve the interpretability of XAI models for drug discovery:

\begin{itemize}
    \item \textit{Model visualization:} XAI models that provide visual representations of their predictions and decisions can help researchers to understand how they are making predictions~\cite{dhanorkar2021needs}. For example, decision trees and rule-based models can be visualized as a tree or set of rules, respectively, which can provide insights into the model's decision-making process.

    \item \textit{Model explanation:} XAI models that provide clear and transparent explanations for their predictions and decisions can help to improve the interpretability of AI algorithms~\cite{adadi2020explainable}. This can be done by providing an overview of the model's decision-making process or by highlighting the most important features that influenced the prediction.

    \item \textit{Feature importance:} XAI models that provide information about the importance of individual features in their predictions can help researchers to understand how the model is using different variables to make decisions~\cite{zhang2021survey}. This can be useful in identifying potential limitations or flaws in the algorithm.

    \item \textit{Model comparison:} Comparing different XAI models and their predictions can help researchers to identify the strengths and weaknesses of each model, and to understand how they make predictions~\cite{hailemariam2020empirical}.

    \item \textit{Human oversight:} Finally, human oversight is crucial in ensuring the interpretability of XAI models. This can involve involving domain experts in the development and validation of XAI models, and regularly reviewing the models' outcomes to identify any limitations or flaws~\cite{langer2021we}.
\end{itemize}

\begin{table*}[!hbtp]
  \begin{center}
    \caption{Contribution of XAI technology for Drug Discovery and its potential application and treatment.}
    \label{tab:nanotab}
    \begin{tabular}{p{1.5cm} p{1.5cm} p{1cm} p{1.5cm} p{1.5cm} p{2cm} p{1.5cm}}
    \hline
    \textbf{Reference} & 
    \textbf{Drug type} &
    \textbf{XAI Model} &
    \textbf{Scheme} &
    \textbf{Actuation} &
    \textbf{Application} &
    \textbf{Treatment} \\
    \hline
       ~\cite{dimitsaki2023benchmarking} & Small molecule & XAI & Bayesian Network & In silico screening & Identifying potential leads for SARS-CoV-2 & COVID-19 \\    \hline
         ~\cite{hinnerichs2021dti} & Protein-based & ML & Decision tree & Target identification & Predicting drug-protein interactions & Cancer \\    \hline
        ~\cite{paul2021artificial} & Small molecule& DL & CNN & Compound optimization & Predicting compound potency & Cardiovascular disease \\    \hline
        ~\cite{ogiya2023identification} & RNA-based & NN & RNN & Target prediction & Identifying drug targets for genetic disorders & Genetic disorders \\    \hline
        ~\cite{vijayan2021enhancing} & Protein-based & XAI rules & Rule-based & Target prediction & Identifying potential drug targets & Neurological disorders \\    \hline
    \end{tabular}
  \end{center}
\end{table*}

\subsection{Developed XAI Modelling Frameworks}
Several frameworks can be used to develop Explainable Artificial Intelligence (XAI) models for drug discovery. Some of the most common frameworks include:

\subsubsection{Decision Trees}
Decision trees are a simple and interpretable XAI framework that can be used for drug discovery. They work by dividing a dataset into smaller subsets based on the values of individual features and making predictions based on the majority class in each subset. Decision trees can be visualized as tree-like structures, which makes it easy to understand the model's decision-making process. The advantage of decision trees is that they are easy to interpret and understand, making them a good choice for XAI models~\cite{gerlach2022decision}. They also tend to be relatively fast to train and can handle both numerical and categorical data, which makes them well-suited for a variety of applications. 

For a drug delivery XAI model, the decision tree could be used to predict the most appropriate delivery method for a patient based on various patient characteristics and delivery-related factors~\cite{chen2021applications}. For example, the tree might consider factors such as patient age, weight, medical history, and the type of drug being delivered. The tree would split the data based on the values of these features and determine the best delivery method for each subset of patients.

However, decision trees are not always the best choice for every XAI problem. They can be prone to overfitting, especially when dealing with a large number of features, and they can become very complex, making them difficult to interpret~\cite{custode2023evolutionary}. In these cases, other machine learning techniques, such as random forests or gradient boosting, might be a better choice.

\subsubsection{Rule-Based Models}
Rule-based models are another simple and interpretable XAI framework that can be used for drug discovery. They work by using a set of rules to make predictions, based on the values of individual features. Rule-based models can be visualized as a set of rules, which makes it easy to understand the model's decision-making process~\cite{vijayan2021enhancing}.

For a drug delivery XAI model, a rule-based system could be used to determine the most appropriate delivery method for a patient based on various patient characteristics and delivery-related factors. For example, the system might include rules such as "if the patient is elderly and has a history of respiratory problems, then use a nebulizer for drug delivery," or "if the patient is younger and has no history of heart problems, then use an intravenous delivery method." The advantage of rule-based models is that they are highly interpretable and provide clear and transparent explanations of how decisions are made~\cite{aghaeipoor2023fuzzy}. This makes them a good choice for XAI models where understanding the reasoning behind decisions is important. They are also easy to develop and maintain, especially for problems where the underlying rules are well-understood~\cite{streeb2021task}.

However, rule-based models can become complex and difficult to manage as the number of rules grows, and they may not be flexible enough to handle all possible scenarios~\cite{nguyen2022attentive}. In these cases, other machine learning techniques, such as decision trees or neural networks, might be a better choice. Additionally, rule-based models may be limited by the quality of the rules that are defined, so it is important to carefully consider the knowledge and expertise of the experts involved in defining the rules.

\subsubsection{Deep Learning Models}
Deep learning models are a more complex XAI framework that can be used for drug discovery. They work by using a deep neural network to make predictions, based on the values of individual features~\cite{kobylinska2022explainable}. Deep learning models can provide accurate predictions, but they can be difficult to interpret. For example, a deep learning XAI model for drug delivery might use a neural network to predict the most appropriate delivery method for a patient based on various patient characteristics and delivery-related factors~\cite{hauser2022explainable}. The model could then use techniques such as activation maximization or saliency maps to explain how the predictions were made, highlighting which input features had the most impact on the prediction.

The advantage of deep learning XAI models is that they can handle complex and non-linear relationships between inputs and outputs, making them well-suited for problems with a large amount of data and complex relationships~\cite{ahmed2021explainable}. They can also learn patterns and relationships in the data that may not be immediately apparent to humans. However, developing XAI models that are both accurate and interpretable can be a challenge. There is often a trade-off between accuracy and interpretability, and it can be difficult to balance the two. Additionally, deep learning models can be difficult to fine-tune and optimize, and a large number of parameters can make them more prone to overfitting the data.

\subsubsection{Hybrid Models}
Hybrid models are a combination of multiple XAI frameworks and can be used to combine the strengths of different models. For example, a hybrid model might use a deep neural network to make predictions but also provide a decision tree as an explanation for the predictions~\cite{sohail2022xai}. The advantage of hybrid XAI models is that they can provide a balance between the interpretability and accuracy of the individual models, making it easier for humans to understand how decisions are made. They can also handle complex and non-linear relationships between inputs and outputs, making them well-suited for problems with a large amount of data and complex relationships~\cite{kaplun2021cancer}.

For example, a hybrid XAI model might combine a rule-based system with a deep learning model. The rule-based system would provide an initial set of predictions based on expert knowledge or experience, and the deep learning model would then refine these predictions based on the relationships it has learned from the data~\cite{kavitha2022deep}. The hybrid model could then provide an explanation of its predictions by combining the explanations provided by the rule-based system and the deep learning model. However, developing hybrid XAI models can be more complex than developing individual models, and careful consideration must be given to how the different models will interact and complement each other. Additionally, there may be a trade-off between interpretability and accuracy, and it can be difficult to balance the two.

Apart from the aforementioned models, several XAI frameworks can be used for drug discovery, each with its own strengths and weaknesses. The choice of the framework will depend on the specific requirements of each drug discovery project, including the size and complexity of the dataset, the interpretability of the model, and the accuracy of the predictions.


\section{Open research challenges and Future Directions}
\label{sec:open}
In the pursuit of advancing XAI within the realm of drug discovery and development, there lie several intriguing avenues that invite exploration. This section delves into the multifaceted landscape of challenges yet to be conquered and promising directions that impress researchers, practitioners, and stakeholders alike. From addressing the challenges of interpretable model complexity to promoting uncharted ethical dimensions, these open challenges and future directions reshape the way we comprehend, utilize, and ultimately transform the field of drug discovery. 

\subsection{Open research challenges}

\subsubsection{Interpretable Model Complexity} 
Developing XAI models that can successfully explain the predictions of sophisticated machine learning models, such as deep learning models, which are frequently employed in drug discovery. Model complexity and interpretability still need to be balanced. It is crucial to provide explanations that dissect the complex decision-making of these advanced models since drug discovery entails complex biological interactions and enormous datasets~\cite{kirbouga2023explainability}. To achieve this balance, novel methods must be developed that transform the complicated models' high-dimensional interactions into understandable insights, enabling researchers and subject-matter experts to understand and believe AI-driven forecasts~\cite{liu2023interpretable}.

\subsubsection{Quantifiable Explanation Metrics} 
Defining and standardizing metrics to tally the effectiveness of XAI-provided justifications. This would make it possible to evaluate these explanations' dependability and informational value objectively. It is crucial to establish reliable measures that measure explanations' ability to reveal innovative insights as well as their alignment with domain knowledge. The credibility and comparability of various XAI techniques will also be improved by creating a unified framework for evaluating these metrics across various drug discovery domains, making it easier to choose and optimize the best techniques for particular research contexts~\cite{kashyap2022quantification, becker2022green}.

\subsubsection{Multi-Modal Data Integration} 
Constructing XAI models that can logically combine and explain predictions drawn from a variety of data sources, including clinical data, omics data, and molecular structures. It is difficult to ensure consistent and comprehensible explanations across these many data kinds. The need for XAI approaches that can integrate knowledge from genomes, proteomics, chemical structures, and patient profiles is critical given the large quantity of data that goes into modern drug discovery~\cite{boehm2022harnessing}. This necessitates not just a knowledge of complex machine-learning techniques but also the capacity to translate these insights into understandable justifications that physicians and researchers from other fields can rely on and act upon. The creation of adaptive XAI frameworks that integrate these various sources into a unified story may hasten the identification of new therapeutic approaches in this multidimensional world~\cite{vermeulen2022multimodal}.

\subsubsection{Trustworthiness and Robustness} 
XAI explanations must be reliable and resilient, particularly when facing adversarial attacks or noisy data. It is essential to create tools that can recognize and counteract false or biased claims. The vulnerability of XAI systems to adversarial manipulation and uncertainty in real-world data creates a significant problem in the dynamic environment of drug development, where trustworthy conclusions are crucial. The confidence of researchers and doctors will increase by fostering a new paradigm of XAI that not only clarifies the AI's reasoning but also protects against potential distortions~\cite{holzinger2022information}. The foundation for informed decision-making will be strengthened by working toward explanations that are not only understandable but also resilient in the face of varied data complexities and possibly malevolent impacts, propelling the ethical and efficient use of AI in the process~\cite{hong2023international}.

\subsubsection{Human-Computer Interaction} 
Designing user-friendly user interfaces that successfully communicate XAI-generated explanations to domain experts in drug discovery, allowing them to base their decisions on these explanations. It's difficult to strike the ideal balance between information richness and simplicity. Constructing user-friendly interfaces that convert complex model outputs into useful insights is a multifaceted task as XAI emerges as a bridge between complex machine insights and human cognition. This covers both the cognitive psychology involved in communicating complicated information as well as the aesthetics of visualization~\cite{kumar2022review}. A critical step toward realizing the transformative potential of AI within the dynamic domain is making sure that these interfaces cater to a variety of user backgrounds, enabling researchers, clinicians, and decision-makers to gain crucial insights from AI-driven predictions while maintaining the depth of technical understanding~\cite{alrizq2022architecture}.

\subsection{Future research directions}

\subsubsection{Dynamic Explanation Adaptation} 
Looking into methods for dynamically changing XAI explanations as models change and new data streams in. This study might produce explanations that continue to be correct, pertinent, and in line with how drug development is evolving~\cite{matsuzaka2022applications}. The insights provided to researchers and clinicians can maintain their interpretability and reliability by using adaptable XAI frameworks that automatically update explanations in response to model updates or changes in data distribution~\cite{bienefeld2023solving}. This will enable informed decision-making even in the face of changing complexity.
	
\subsubsection{Ethical and Societal Implications} 
Exploring the ethical dimensions of XAI in drug development, such as questions of prejudice, justice, openness, and data privacy. It will be essential for the ethical implementation of AI-driven healthcare solutions to comprehend and handle these ethical issues. We can proactively build systems to eliminate biases, maintain openness, and safeguard patient data, creating trust and responsibility within the healthcare ecosystem, by investigating the ethical implications of utilizing AI to drive important decisions in drug development~\cite{jorg2023medaicine, albahri2023systematic}.

\subsubsection{Hybrid Models and Fusion Strategies} 
Study of hybrid models, which combine the benefits of comprehensible methods with the propensity for prognosis of complex models. Creating fusion tactics that seamlessly combine these methods could result in explanations that are precise and comprehensible. We can combine the benefits of sophisticated machine learning and interpretable models to develop a new class of hybrid models that provide predictions with high accuracy and accompanying insights that are simple to understand, facilitating sound decision-making~\cite{kirbouga2023explainability, albahri2023systematic}.

\subsubsection{Longitudinal Data and Temporal Insights} 
XAI techniques that can explain predictions based on patient longitudinal data, capturing illness development and treatment responses across time, are the focus of this study. Personalized treatment plans might undergo a revolution if interpretable insights into temporal dynamics are made possible~\cite{anguita2020explainable}. XAI techniques that can understand the complex temporal relationships within these data streams will enable clinicians to identify crucial disease milestones, evaluate treatment efficacy, and design interventions that adapt to the changing state of patient health as healthcare data becomes more longitudinal and dynamic~\cite{chen2021applications}.
	
\subsubsection{Domain-Specific Explanation Languages} 
Designing domain-specific languages or frameworks that make it easier for multiple stakeholders, like as regulators, patients, and regulatory agencies, to translate complex AI-generated explanations into insights they can use~\cite{brynjulfsen2021xai}. We can bridge the gap between technical AI outputs and practical decision-making by adapting the language of explanations to the unique needs and backgrounds of varied users~\cite{thirunavukarasu2023large}. This will allow stakeholders at all levels to effectively utilize the potential of AI-driven insights in drug discovery.

\section{Conclusions} 
\label{sec:conclusion}
The landscape of drug discovery has undergone a profound shift with the integration of AI and ML. As these technologies advance, the demand for transparency and interpretability has become increasingly paramount. XAI is a groundbreaking approach that tackles this very concern by offering a comprehensible insight into the outcomes produced by intricate ML models. The presented review article has presented a comprehensive spotlight on the current status of XAI within drug discovery. Encompassing a range of XAI methods and their application in this domain, the article underscores both the potential and the challenges of implementing XAI techniques. The examination extends to XAI's role in diverse drug discovery facets, such as target identification, compound design, and toxicity prediction. With a forward-looking perspective, the article also contemplates potential research trajectories, envisioning how XAI could further reshape the landscape of drug discovery. Ultimately, this review serves as a valuable compass, guiding us through the evolving terrain of XAI's transformative impact on the field of drug discovery.


\printcredits

\bibliographystyle{unsrtnat}


\bibliography{References}
\end{document}